\documentclass{article}

\usepackage[accepted]{icml2020}

\icmltitlerunning{Circuit-Based Intrinsic Methods to Detect Overfitting}

\usepackage[utf8]{inputenc} 
\usepackage[T1]{fontenc}    
\usepackage{hyperref}       
\usepackage{url}            
\usepackage{booktabs}       
\usepackage{amsfonts}       
\usepackage{nicefrac}       
\usepackage{microtype}      
\usepackage[pdftex]{graphicx}
\usepackage{subcaption}
\usepackage{balance}

\begin{document}

\twocolumn[
\icmltitle{Circuit-Based Intrinsic Methods to Detect Overfitting}



\icmlsetsymbol{equal}{*}

\begin{icmlauthorlist}
\icmlauthor{Satrajit Chatterjee}{google}
\icmlauthor{Alan Mishchenko}{berkeley}
\end{icmlauthorlist}

\icmlaffiliation{berkeley}{Department of EECS, University of California, Berkeley, California, USA}
\icmlaffiliation{google}{Google, Mountain View, California, USA}

\icmlcorrespondingauthor{Satrajit Chatterjee}{schatter@google.com}
\icmlcorrespondingauthor{Alan Mishchenko}{alanmi@berkeley.edu}

\icmlkeywords{Machine Learning, generalization, neural networks, overfitting, memorization, random forests}

\vskip 0.3in
]

\printAffiliationsAndNotice{}

\begin{abstract}

The focus of this paper is on intrinsic methods to detect overfitting.  By {\em intrinsic} methods, we mean methods that
    rely only on the model and the training data, as opposed to traditional
    methods (we call them {\em extrinsic} methods) that rely on performance on a test set or on bounds from
    model complexity.
We propose a family of intrinsic methods called Counterfactual Simulation (CFS)
    which analyze the flow of training examples through the model by
    identifying and perturbing rare patterns. By applying CFS to logic circuits
    we get a method that has no hyper-parameters and works uniformly across
    different types of models such as neural networks, random forests and
    lookup tables.
Experimentally, CFS can separate models with different levels of overfit using
    only their logic circuit representations without any access to the high
    level structure.  By comparing lookup tables, neural networks, and random
    forests using CFS, we get insight into why neural networks generalize. In
    particular, we find that stochastic gradient descent in neural nets does
    not lead to ``brute force'' memorization, but finds common patterns (whether
    we train with actual or randomized labels), and neural networks are not
    unlike forests in this regard.
Finally, we identify a limitation with our proposal that makes it unsuitable
    in an adversarial setting, but points the way to future work on robust
    intrinsic methods.

\end{abstract}

\section{Introduction}
\label{intro}

This paper considers methods to detect overfitting of a model 
based only on the model and the training data.  In terminology that we introduce, we call such methods 
{\em intrinsic}, in contrast to {\em extrinsic}
methods, which rely on additional knowledge, such as, the performance of the
model on examples held out from the training process, details of the process
used to find the model (e.g., multiple hypothesis testing with registration), or
limitations of the function family to which the model belongs (e.g., VC
dimension, Rademacher complexity) or those of the size of the parameter space of the
model (e.g., Akaike Information Criterion). 

We classify methods relying on the knowledge of the function family as extrinsic because often the information relevant to overfitting is not directly represented in a given model. For example, a model could have been found by searching a much smaller space than that implied na\"ively by the function family to which it belongs, due to either explicit regularization or the regularization implicit in the optimization or search procedure. Conversely, the given model could have been found by searching a much larger space of models through hyper-parameter search, or by picking the best model from a number of different model families, but the specific final model itself does not carry any vestiges of the larger space that was searched over. Both of these situations are common in modern machine learning.  

Intrinsic methods are of practical interest since
with modern deep learning models, we find that extrinsic estimates based on model
complexity are typically vacuous since these models are powerful enough to fit
arbitrary data~\citep{Zhang17}.
Consequently, practitioners resort to studying performance on a hold out
dataset (or cross validation), but this is unsatisfactory for a couple of
reasons. First, this means that, in a low data setting, we cannot use all the
data for training, but have to keep significant portions aside for validation
(e.g., see discussion in \citet{Dietterich98}).
Second, it may be difficult to ensure a pristine hold out that is not touched
during the research process particularly if the project is long running. Even
with a few queries to the hold out during the research process, it is possible
to start fitting to the hold out~\citep{Dwork15}.

Intrinsic methods are also interesting from a theoretical perspective. Imagine
that we have sufficient computing power to, say, enumerate all neural networks
(and their weights) up to a certain size. Among all the networks that fit the
data well, intrinsic methods could distinguish between those networks that
generalize well from those that do not, and we could view the model as a {\em
certificate} of generalization.\footnote{Keeping a hold out set would not help
us here---what would that even mean?} In addition, if the intrinsic method was
efficient, it would mean that supervised learning (and not just fitting the training data) is in
{\sf NP}. Intrinsic methods can also help shed light on why neural networks
trained with stochastic gradient descent generalize in spite of their large
capacity. Some recent analyses based on normalized margin, curvature, etc.~\citep{Bartlett17, Rangamani19, Arora18, Neyshabur18}
may be seen as intrinsic estimates for generalization albeit specialized
to neural networks.

Intrinsic methods can be considered in the context of a protocol involving
two agents. Let $\mathcal{S}$ be a public dataset drawn from a distribution
$\mathcal{D}$ that generates samples infrequently (e.g.,  quarterly financial
statements and stock market returns of public companies, or public health data on treatments
and patient outcomes). Suppose Arthur wants to build a model from $\mathcal{S}$
but instead of doing so himself, he outsources it to Merlin, an untrusted
adversary.  Merlin comes back with a model $\mathcal{M}$ but does not disclose
any details of his modeling process. How can Arthur convince himself that
$\mathcal{M}$ is not horribly overfit? For example, $\mathcal{M}$ could simply
be a lookup table built from $\mathcal{S}$.  Normally Arthur would evaluate
$\mathcal{M}$ on new samples from $\mathcal{D}$, but in our setup, Arthur does
not have any samples other than those in $\mathcal{S}$ since all the existing
data is public.
Now, if Arthur only has access to $\mathcal{M}$ as a black box and he can only
evaluate $\mathcal{M}$ on elements of $\mathcal{S}$, it appears there is little
he can do to distinguish a good model from a lookup table. But {\em can Arthur 
do better if he has access to the internal signals in the implementation of 
$\mathcal{M}$?} This is the central question of this paper.
 
We take a first step towards answering this question by studying a
naturally-motivated family of intrinsic methods, called {\em Counterfactual
Simulation} (CFS), and evaluating their efficacy experimentally on a benchmark
problem. The main idea behind CFS is to analyze the flow of the training
examples in $\mathcal{S}$ through the structure of $\mathcal{M}$.
This is only a first step since CFS, although promising in practice, has
significant limitations. In particular, our experiments show that even if we
could prove bounds based on CFS, they would not be tight enough in an
adversarial setting. However, we hope that this paper encourages research to
overcome these limitations or to show that no such method
can exist, especially for learning tasks of practical interest.

\begin{figure}[t]
\begin{center}
\vskip 0.2in
\centerline{\includegraphics[width=\linewidth]{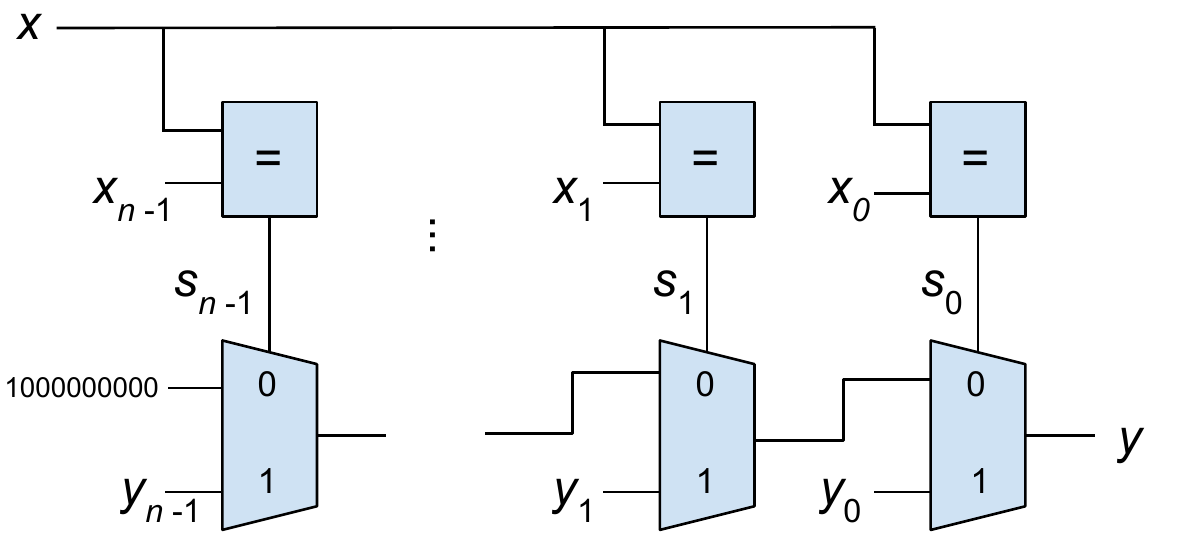}}
\vskip 0.2in
    \caption{A circuit implementing a lookup table that memorizes the training examples $(x_i, y_i)$ for $0 \le i < n$. We observe that in this extreme case of overfitting, there are signals in the circuit (the $s_i$) that identify specific
    training examples. For e.g., $s_0$ is 1 only when $x = x_0$ and 0 for all other $x_i$ (assuming the $x_i$ are distinct). 
    We say that 1 is a {\em rare pattern} for the signal $s_0$. 
    Based on this example, we propose that the occurrence of rare patterns during simulation of a training set through a model indicates overfitting, and 
    in this work, we explore to what extent such rare patterns can be used to detect overfitting in more complex models such as neural networks and random forests.} 
\label{fig:lut_circuit}
\end{center}
\vskip -0.2in
\end{figure}

\section{Counterfactual Simulation (CFS)}

The structure of $\mathcal{M}$ can be described at different levels of
abstraction.  For instance, if $\mathcal{M}$ is a fully connected feed forward
neural network, we can describe it as a sequence of layers.  Going one level
lower, we can describe it as a directed acyclic graph (DAG) of fixed or
floating point adders, multipliers, and pointwise non-linearities. Finally, at
the lowest level of abstraction, we can describe the structure of $\mathcal{M}$
as a DAG of primitive logic gates such as 2-input And gates and inverters,
i.e., as a combinational logic circuit.
In our setup, it is natural to work at this lowest level, i.e.,
logic gates since it allows different kinds of models such as lookup tables,
random forests, and neural networks all to be mapped into the same format.
Thus, Merlin need not disclose even what type of model he has built, but simply
provides Arthur with a combinational logic circuit for the model. 

To make this concrete, consider the MNIST image classification problem~\citep{lecun10} which we
will use as a running example. $\mathcal{D}$ is the distribution of handwritten
digits and their classes. $\mathcal{S}$ is a sample from $\mathcal{D}$ of
60,000 images $x_i$ and their corresponding labels $y_i$ (thus, 0 $\le$ $i$ $<$
$60000$) i.e., the MNIST training set. Each $x_i$ is 6,272 bits wide
(corresponding to 28 $\times$ 28 pixels $\times$ 8 bits per pixel), and each
$y_i$ is 10 bits wide (for a 1-hot representation of the 10 possible classes).
Therefore, a classifier to solve this problem is a circuit with 6,272 Boolean
inputs and 10 Boolean outputs.

Suppose Merlin's model for the MNIST classifier is a simple lookup table.  How
would the circuit for it look? Figure~\ref{fig:lut_circuit} sketches one
possibility.  The 6,272-bit input $x$ is compared with each of the examples
$x_i$ in turn and if there is a match, the corresponding 10-bit output $y_i$ is
selected. If no example matches, then the model (arbitrarily) returns the 1-hot
vector representing class `0'.
Now, if we simulate this circuit on examples in $\mathcal{S}$, we notice that
there are internal signals in the circuit that are capable of identifying
specific training examples.  For example, the signal $s_0$ (the output of the
$x = x_0$ block) is 1 (true) for the training example $x_0$ and 0 (false) for
all others. In this case, we say 1 is a {\em rare pattern} for $s_0$ since
$s_0$ rarely takes on the value 1 on the training set. 
Formally, if a signal $s$ in $\mathcal{M}$ takes on the value $v$ at most $l$
times on the training set $\mathcal{S}$, we call $v$ an \underline{$l$-rare pattern} for
$s$.

This observation leads to the first of the two main ideas behind CFS: {\em The
presence of $l$-rare patterns suggests overfitting}, and, therefore, poor
generalization since they open up the possibility that $\mathcal{M}$ has
special logic to detect and handle specific examples.
A count of rare patterns, however, does not directly translate into a metric for
generalization (without building a predictive model of {\em that}!).
Furthermore, although a pattern may be rare it may also be an {\em
observability don't care} (ODC), i.e., it may have no influence on the output
of the circuit.  For example, if the signal with the rare pattern only feeds
into an And gate whose other input is 0 when the rare pattern appears, then the
value of the rare pattern does not matter in deciding the output of the
circuit.

We address both problems with the second main idea behind CFS: {\em perturbed
simulation of a training example} where we simulate an example through $\mathcal{M}$ as usual,
but when we encounter a $l$-rare pattern, instead of propagating it to the fanouts (i.e.,
gates that depend on this signal), we perturb the pattern and simulate the fanouts
with the perturbed pattern.
A natural perturbation is to propagate the opposite value instead of the rare
pattern. 
In our running example, this corresponds to propagating a 0 instead of
1 for signal $s_0$ to the multiplexer when simulating the training image $x_0$.  In
this manner, we prevent the model from identifying $x_0$ and we see that the
output for $x_0$ is no longer necessarily $y_0$. 
We call this modified simulation procedure $l$-counterfactual simulation or \underline{$l$-CFS} for short.

We perform perturbed simulation for each training example in turn and measure
the resulting average accuracy over the training set. We call this quantity the
training accuracy obtained through $l$-CFS.  In our running example of the
lookup table, it is easy to see that the training accuracy obtained through
1-CFS is no better than random chance (since each training example is mapped to
class `0' under 1-CFS). 
Now, since random chance is what one would expect to be the generalization of
the lookup table (i.e., its accuracy on $\mathcal{D}$), it is tempting to
conjecture that 1-CFS training accuracy is a good estimate of accuracy on
$\mathcal{D}$.  Although that is not the case as we shall see empirically in
Section~\ref{sec:expt}, we find that the difference in training accuracy
between normal simulation and $l$-CFS is a good measure of the degree of
overfit of $\mathcal{M}$.

{\bf Other Types of CFS.} There are other variants of the procedure
described above (which we call {\em Simple CFS} or just CFS). Of particular
interest is {\em Composite CFS} which is useful for circuits with gates that have many inputs
or are at higher levels of abstraction. In Composite CFS, we look at
rare patterns in combinations of signals feeding a particular gate and perturb the output of that gate (by flipping it) when a
rare combination is seen at the inputs.  Another possibility is to randomize the
perturbation instead of always flipping.  
In our experiments, we found these variants to produce results that are similar to those obtained from Simple CFS, and so we only mention them in passing.

\section{Experimental Results}
\label{sec:expt}

{\bf CFS Implementation.} Our implementation of $l$-CFS works on a directed acyclic
graph $\mathcal{G}$ representing a combinational logic circuit where each
node is either the constant 0, a primary input, a 2-input And gate, or a
2-input Xor gate. An edge is either a direct connection or an inverter and
represents a Boolean function in terms of the primary inputs. This is a variant of an
And-Inverter Graph~\citep{Biere07, Chatterjee07}, a standard data structure in modern logic synthesis
used to handle circuits with hundreds of millions of nodes. While constructing these graphs, we propagate constants 
but do not extract common sub-expressions.
 
We make two passes through the nodes of $\mathcal{G}$ in topological order
starting from primary inputs.  In the first pass, we simulate the
training set through $\mathcal{G}$ to obtain the counts of different patterns
in the circuit. In the second pass, we use the counts from the first pass to
perturb the $l$-rare patterns. In Simple CFS, this boils down to replacing
signals that take on a value of 0 on most examples with the constant 0 signal
and likewise for 1. CFS thus runs in linear time in the size of the graph
and the training data.
For performance, the simulations are done in a bit parallel manner for all
training examples at the same time. To avoid running out of memory, we use
reference counting to recycle storage for intermediate simulated values when
they are no longer needed. A typical run of $l$-CFS in our experiments takes
less than 10 minutes on a 3.7GHz Xeon CPU and less than 2GB of RAM.

{\bf Benchmark Problem.} While the discussion from the previous section shows
how CFS can discover overfit when the model is a simple lookup table, it is not
clear if CFS would be effective on neural networks trained with stochastic gradient descent (SGD). 
To answer this question, we trained 3 neural networks for MNIST in TensorFlow (Keras) and compiled
them down into combinational logic circuits. All 3 networks have the same
architecture: an input layer of size 784 (i.e., 28 $\times$ 28), 3 fully connected ReLU
layers with 256 nodes each, and a final softmax layer with 10 outputs. (Thus, the
total number of trainable parameters is 335,114.) We also performed some experiments with Fashion MNIST~\citep{Xiao17} 
and the results are similar.

The first two networks ({\tt nn-real-2} and {\tt nn-real-100}), were trained on the
MNIST training set for 2 epochs and 100 epochs respectively. They get to
training (top-1) accuracies of 97\% and 99.90\% respectively.  The third
network ({\tt nn-random}) was trained on a variant of MNIST where the output
labels in the training set are permuted pseudo-randomly and trained for 300
epochs to get to a training accuracy of 91.27\%.\footnote{While evaluating {\tt
random} for training accuracy (with or without CFS), the permuted labels are used.} 
In all cases, we used the ADAM optimizer with default parameters and batch
size of 64. As expected, {\tt nn-real-2} is the least overfit and gets to a
validation set accuracy of 97\% (i.e., has a negligible generalization gap),
{\tt nn-real-100} is more overfit getting to a validation set accuracy of 98.24\%
(a gap of 1.66\%), and finally, the validation accuracy of {\tt nn-random} is
9.73\% (i.e., close to chance) confirming that it is horribly overfit.

{\bf Conversion to Logic Circuits.} This is done by
generating logic subcircuits composed of 2-input And or Xor gates and inverters
for each of the operations in the neural network.  Weights
and activations are represented by signed 8-bit and 16-bit fixed point numbers
respectively with 6 bits reserved for the fractional part. (Weights from
training are clamped to $[-2.0, 2.0)$ before conversion to fixed point.) Each
multiply-accumulate unit multiplies an 8-bit constant (the weight) with a
16-bit input (the activation) and accumulates in 24 bits with saturation. The
constant multiplications are done by finding a minimal combination of
bit-shifts (multiplications by powers of 2) and additions or subtractions. For
example, $5 \times u$ is implemented as $4 \times u + u$ and $11 \times u$ as
$16 \times u - 4 \times u - u$.  ReLUs are implemented with a comparator and a
multiplexer.  The outputs of each network are the 10 signed 16-bit activations before
the softmax. When evaluating accuracy (with CFS or without) we pick the class
corresponding to the largest of the 10 activations (top-1 accuracy).  
The resulting logic circuits have 35 to 52 million And or Xor gates and
5500 to 6000 logic levels. 
(These sizes along with the need to fit random data dictated the choice of
architecture and benchmark.)

\begin{figure*}[p]
\begin{center}

\begin{subfigure}[b]{0.40\textwidth}
    \includegraphics[width=\textwidth]{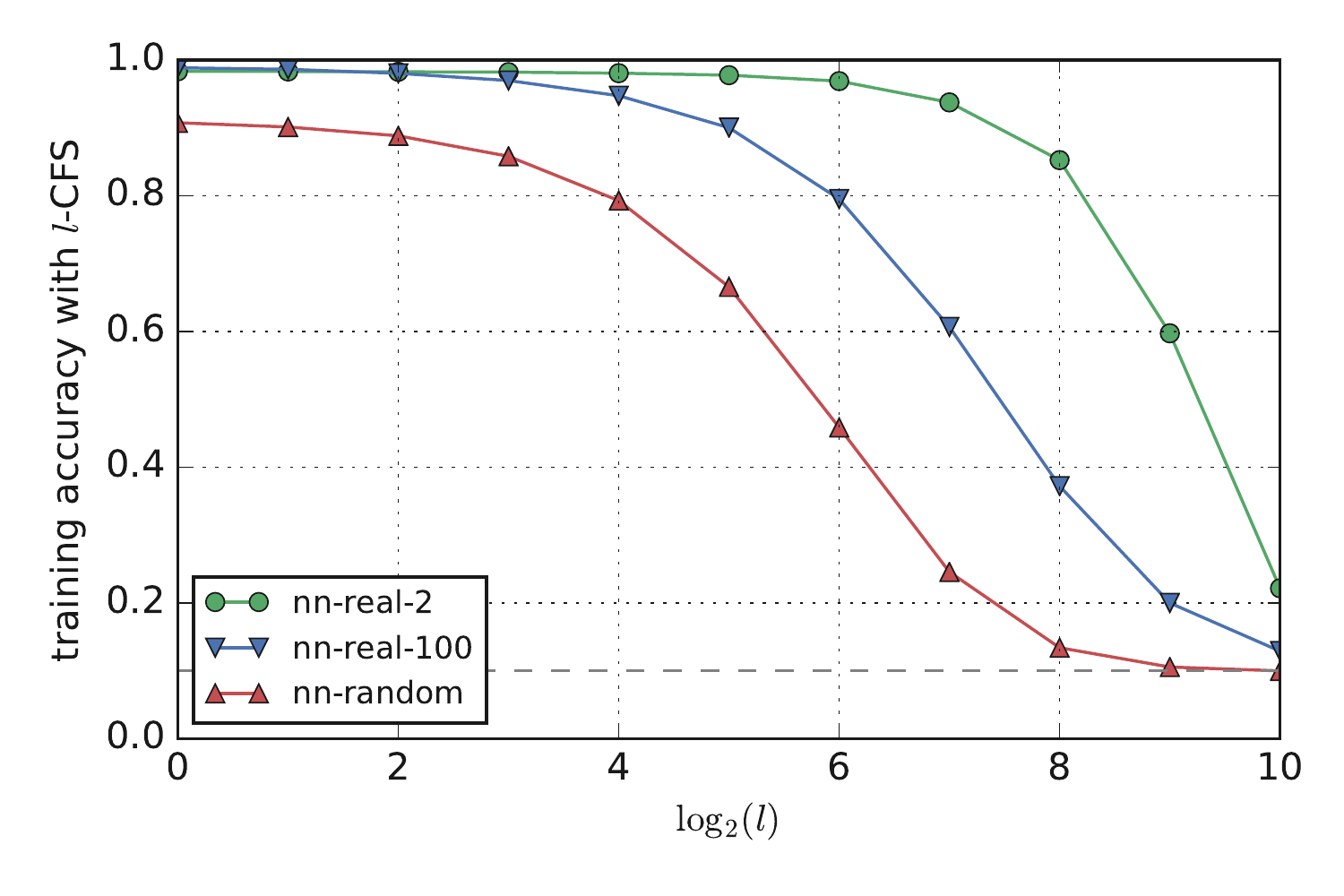}
    \vskip -0.1in
    \caption{}
    \label{fig:expt-1}
\end{subfigure}
    \hspace{1cm}
~ 
\begin{subfigure}[b]{0.40\textwidth}
    \includegraphics[width=\textwidth]{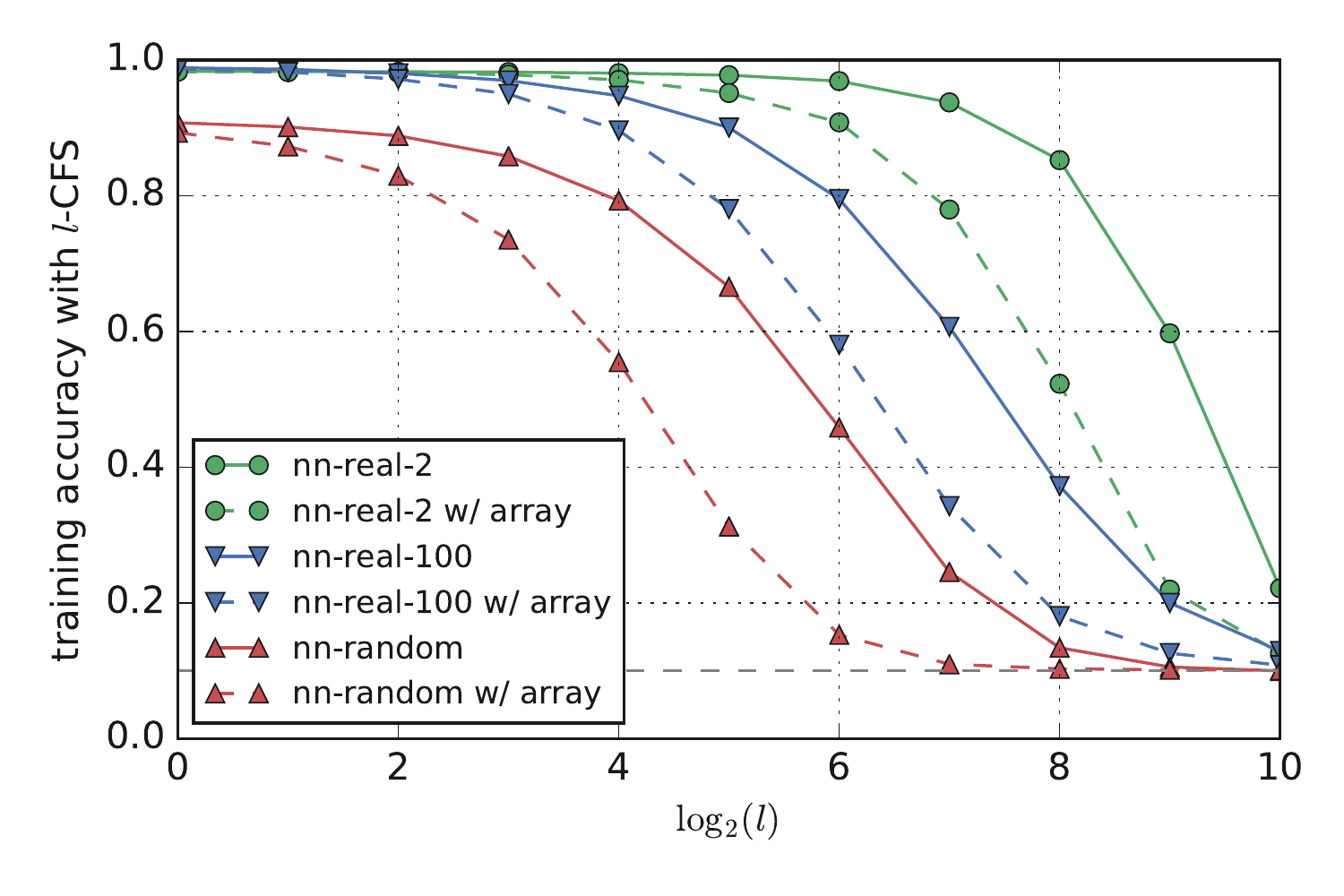}
    \vskip -0.1in
    \caption{}
    \label{fig:expt-2}
\end{subfigure}
 
\begin{subfigure}[b]{0.40\textwidth}
    \includegraphics[width=\textwidth]{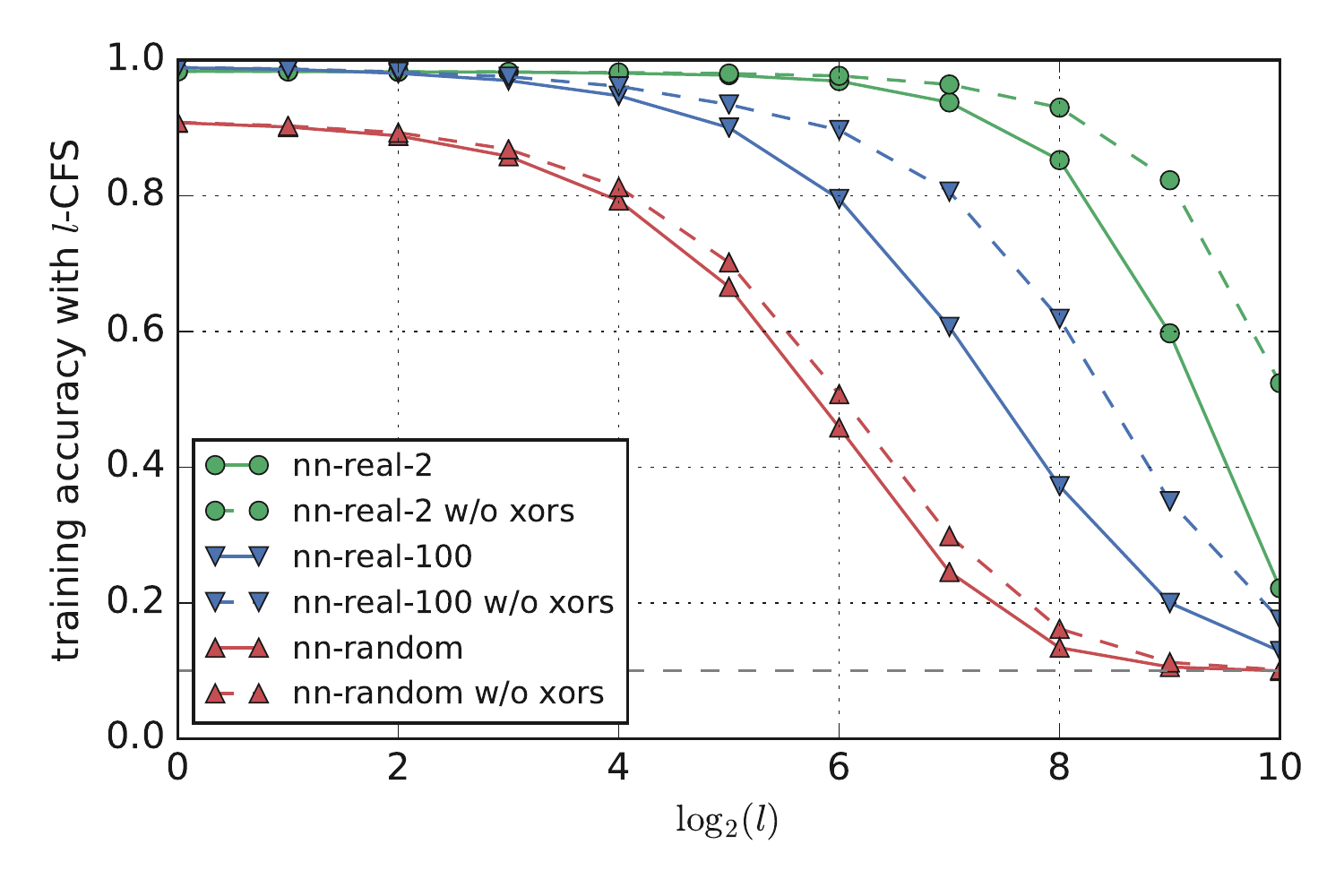}
    \vskip -0.1in
    \caption{}
    \label{fig:expt-3}
\end{subfigure}
    \hspace{1cm}
~
\begin{subfigure}[b]{0.40\textwidth}
    \includegraphics[width=\textwidth]{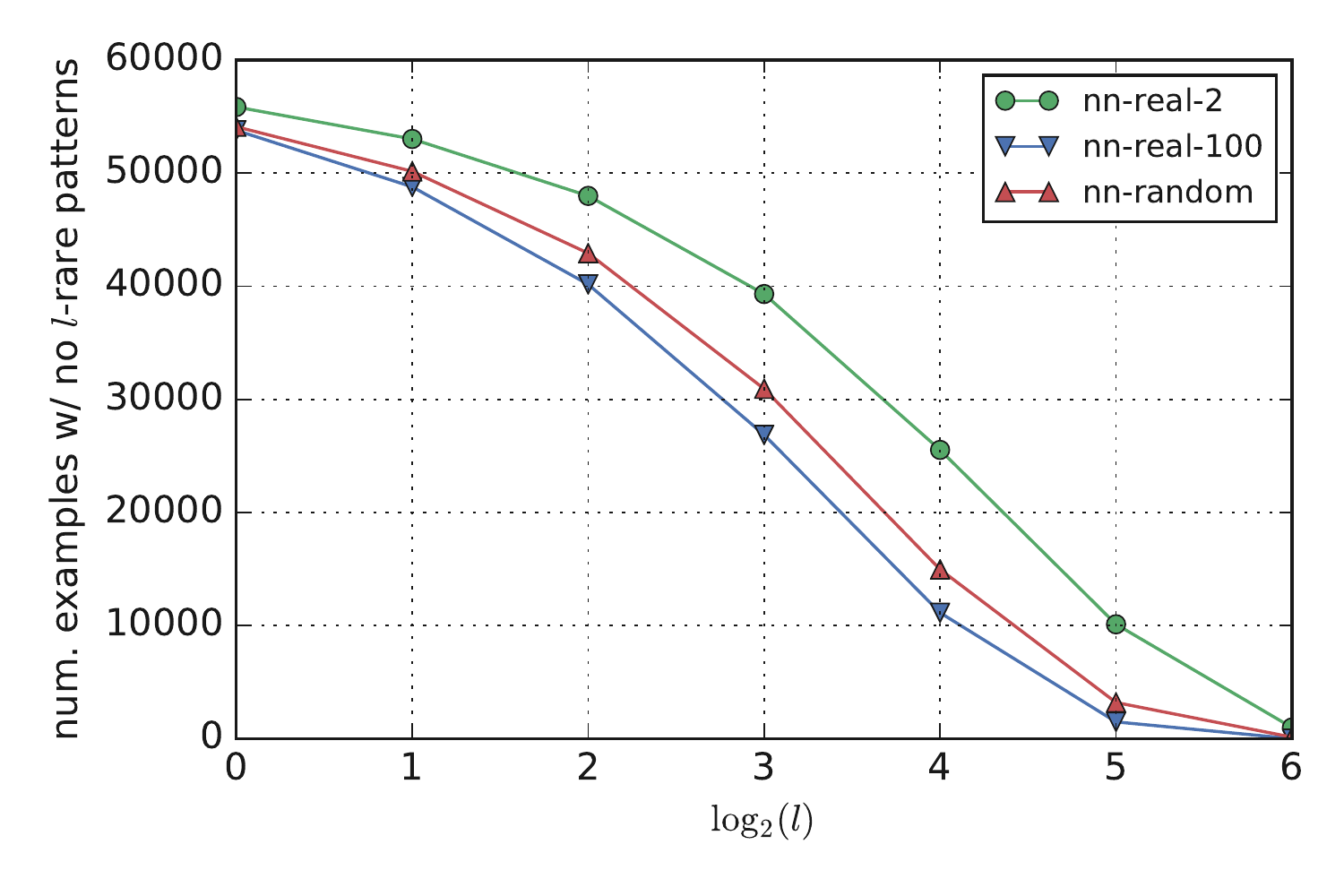}
    \caption{}
    \label{fig:expt-4}
\end{subfigure}
 
\begin{subfigure}[b]{0.40\textwidth}
    \includegraphics[width=\textwidth]{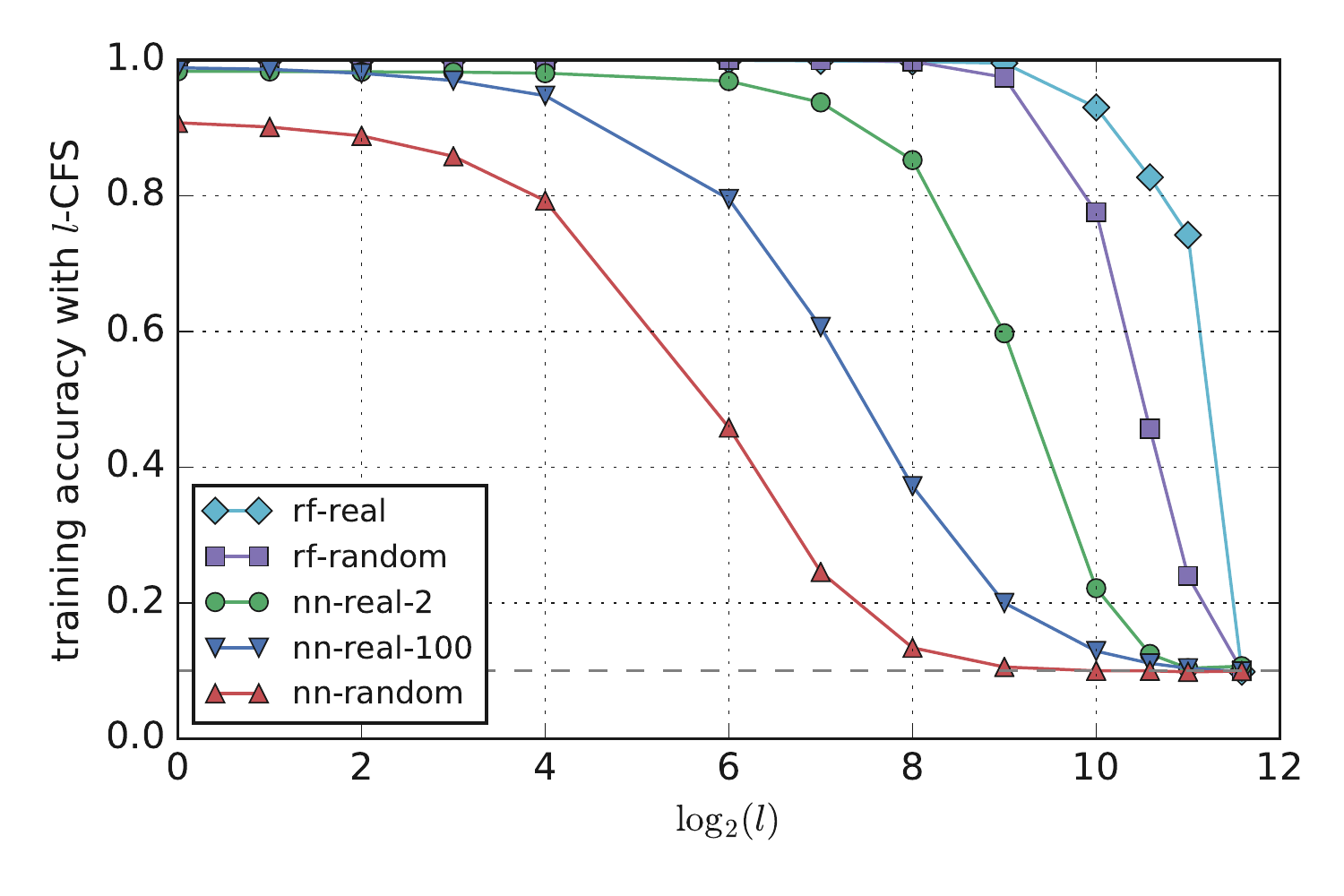}
    \vskip -0.1in
    \caption{}
    \label{fig:expt-5}
\end{subfigure}
    \hspace{1cm}
~
\begin{subfigure}[b]{0.40\textwidth}
    \includegraphics[width=\textwidth]{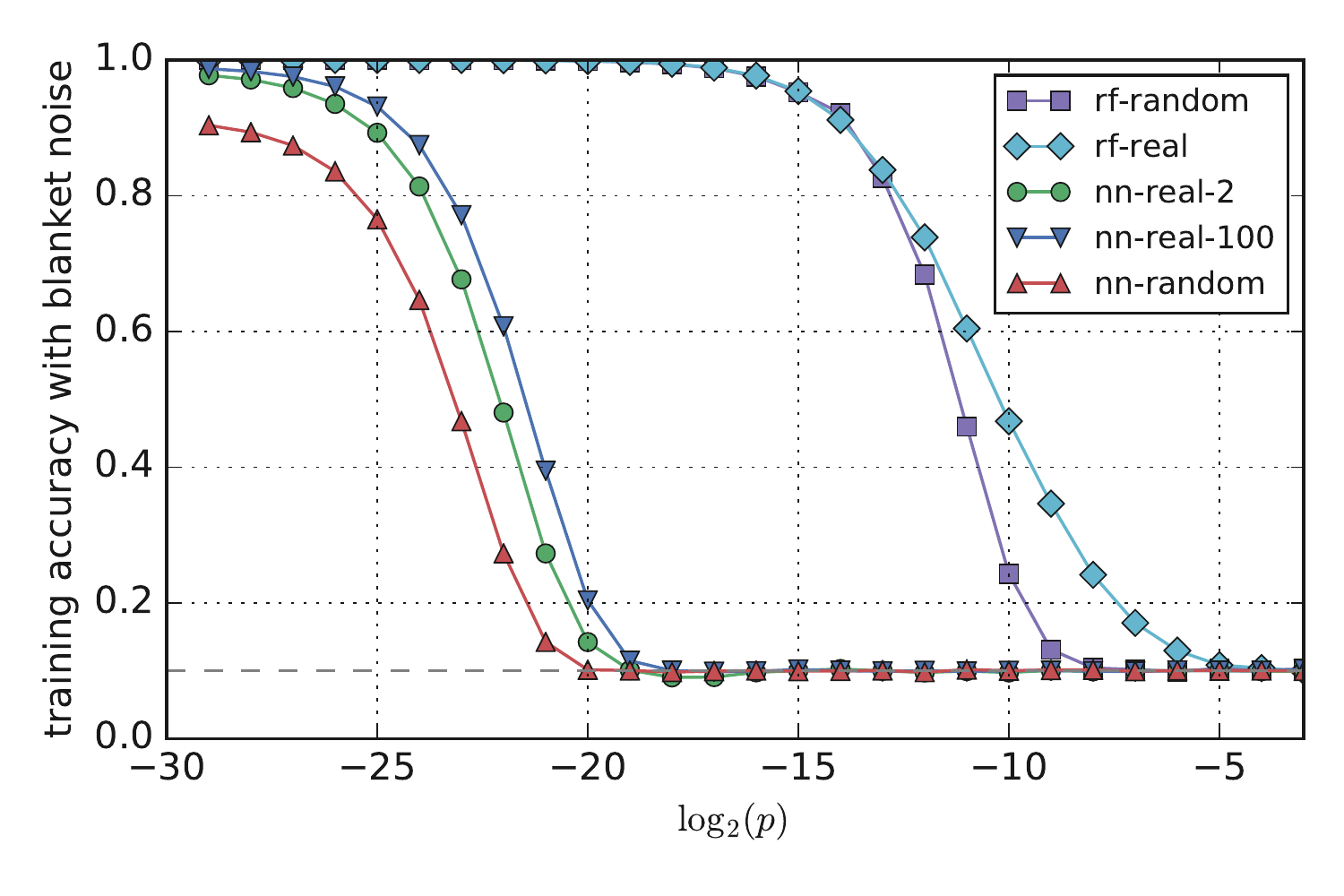}
    \vskip -0.1in
    \caption{}
    \label{fig:expt-6}
\end{subfigure}
 
\begin{subfigure}[b]{0.40\textwidth}
    \includegraphics[width=\textwidth]{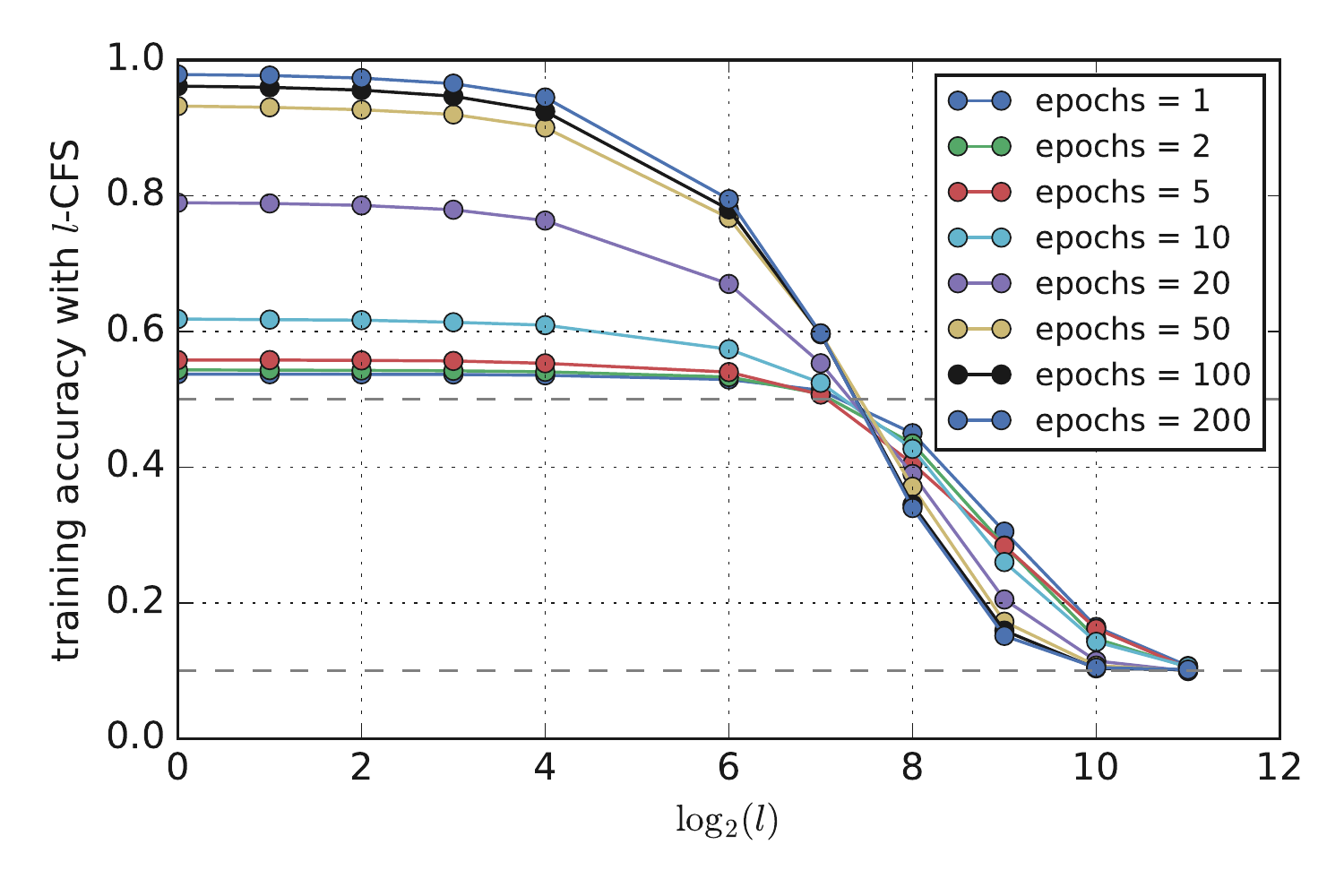}
    \vskip -0.1in
    \caption{}
    \label{fig:expt-7}
\end{subfigure}
    \hspace{1cm}
~
\begin{subfigure}[b]{0.40\textwidth}
    \includegraphics[width=\textwidth]{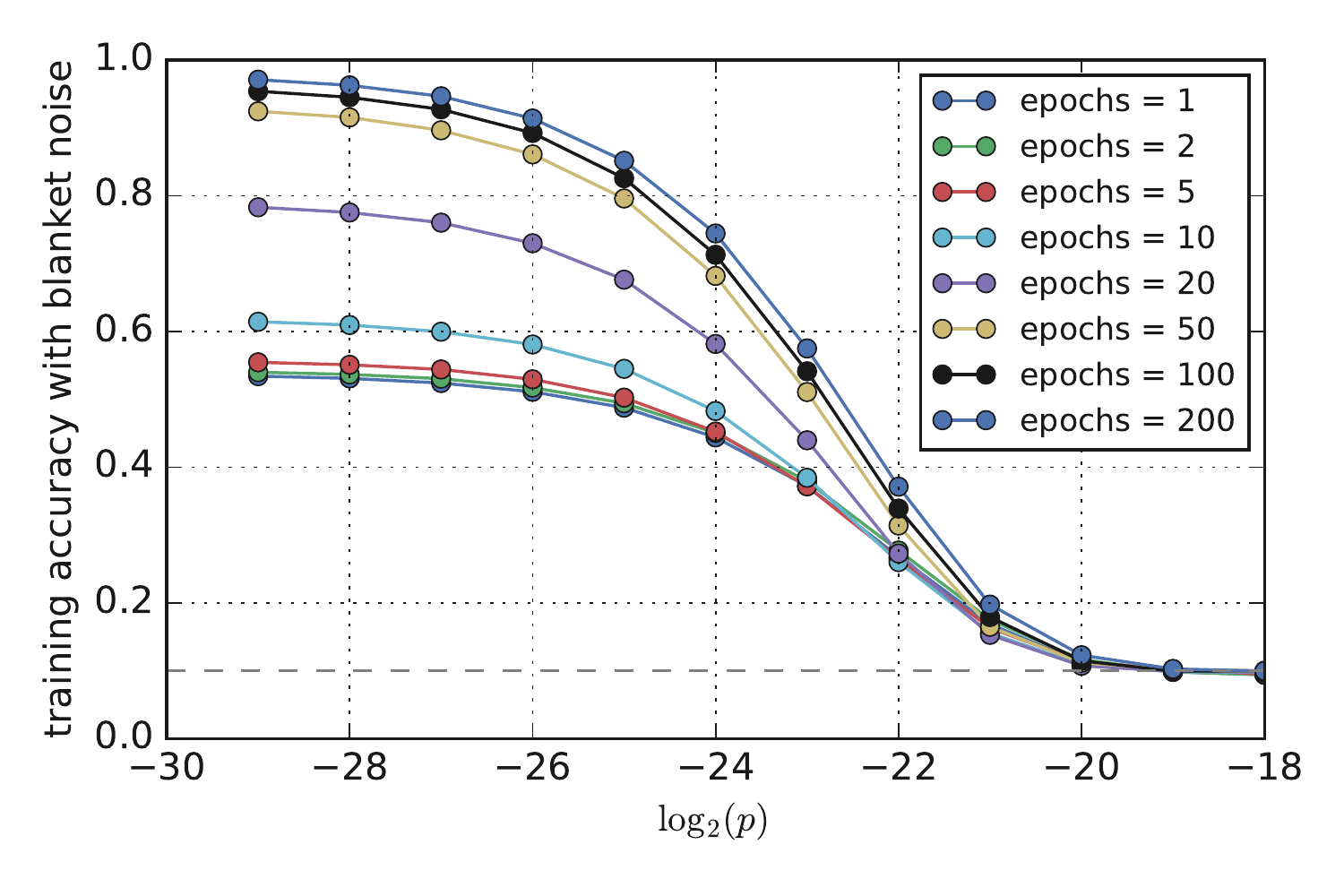}
    \vskip -0.1in
    \caption{}
    \label{fig:expt-8}
\end{subfigure}

    \caption{The results of experiments in Section~\ref{sec:expt}. Plot (a)
    shows the CFS curves for 3 networks with different amounts of overfitting;
    (b) and (c) show the impact of the choice of multipliers and of
    primitive logic gates respectively on CFS curves; (d) shows how many
    examples are unaffected by $l$-CFS since they do not have any $l$-rare
    patterns; (e) shows CFS curves for random forests; (f) shows training
    accuracies when a signal is randomly flipped with probability $p$; and (g)
    and (h) show the differences between CFS and random flips respectively for 8 networks trained 
    on a dataset with high label noise.}

\end{center}
\end{figure*}

{\bf Expt. 1: Effect of Simple CFS.} Figure~\ref{fig:expt-1} shows the training
accuracies obtained through $l$-CFS for each of the three networks as $l$
varies from 1 to 1024 (which is about 1.7\% of the number of training
examples). We call these plots {\em CFS curves}. As $l$ increases, i.e., as
more patterns become rare and get perturbed, the accuracy falls eventually
reaching chance.  However, it is interesting that the drop in accuracy is
highest for {\tt nn-random} (e.g., at $l$ = 64, the drop is about 45\%), somewhat
less for {\tt nn-real-100} (20\%) and the least for {\tt nn-real-2} (1.4\%).  Thus
the falloff in accuracy with $l$ is an indicator of the level of overfit of a
network. 

It is remarkable that even when a neural network is represented 
at a very low level as a logic circuit, relative overfit can be detected
using an intuitive algorithm with no hyper-parameters to tune.

\begin{figure}[!b]
\vskip 0.2in
\begin{center}
\includegraphics[width=0.40\textwidth]{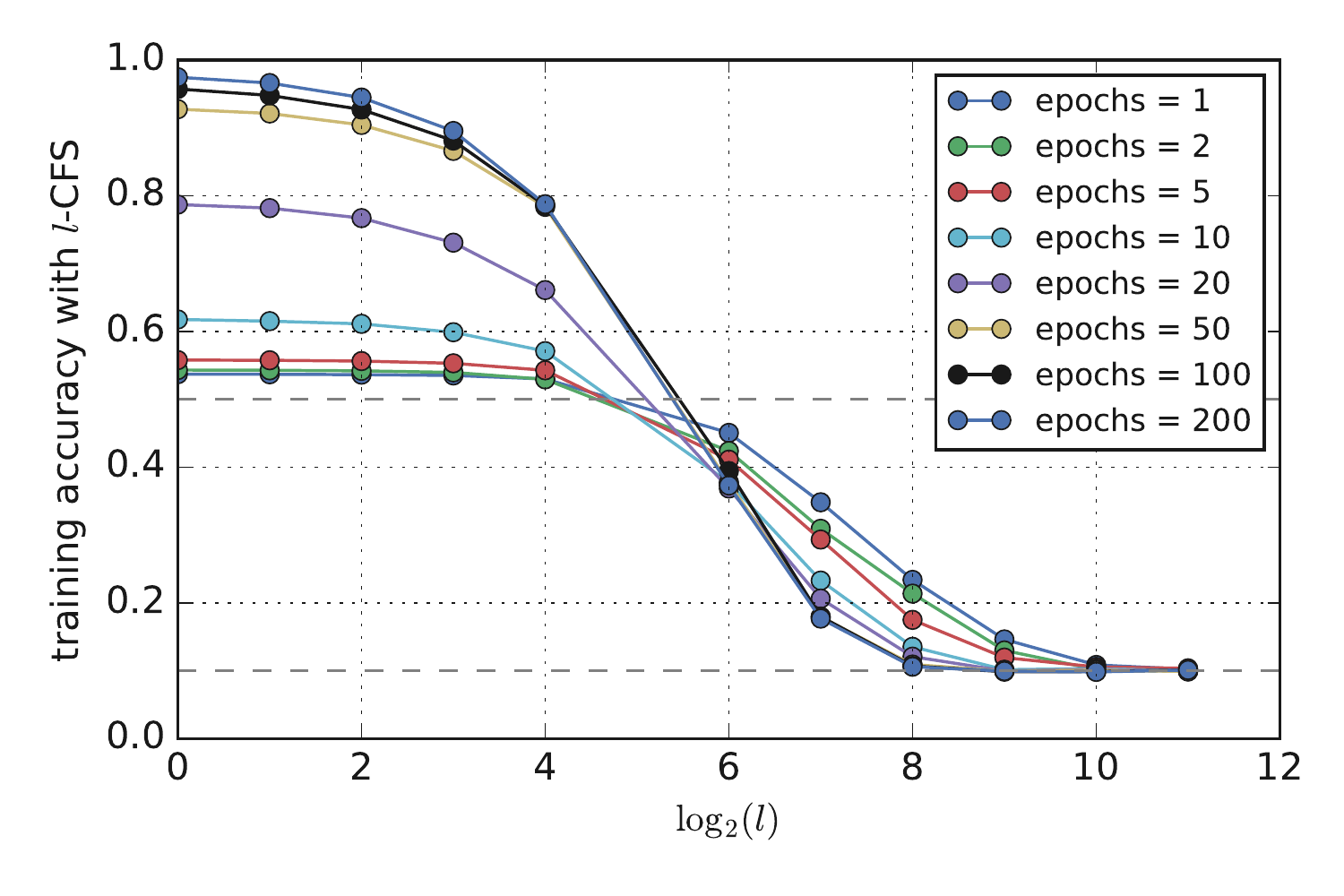}
\caption{
This plot shows the result of repeating Expt. 7 but using Composite CFS (instead of Simple CFS) where we compute and perturb rare patterns across all the signals that feed into a logic gate, i.e., across all the fanins of a gate (instead of computing and perturbing rare patterns for each signal or fanin independently). The results with Composite CFS are very similar to those of Simple CFS (Figure~\ref{fig:expt-7}).
}
\label{fig:expt-composite}
\end{center}
\end{figure}

\begin{figure*}[t]
\vskip 0.2in
\begin{center}

\begin{subfigure}[b]{0.40\textwidth}
    \includegraphics[width=\textwidth]{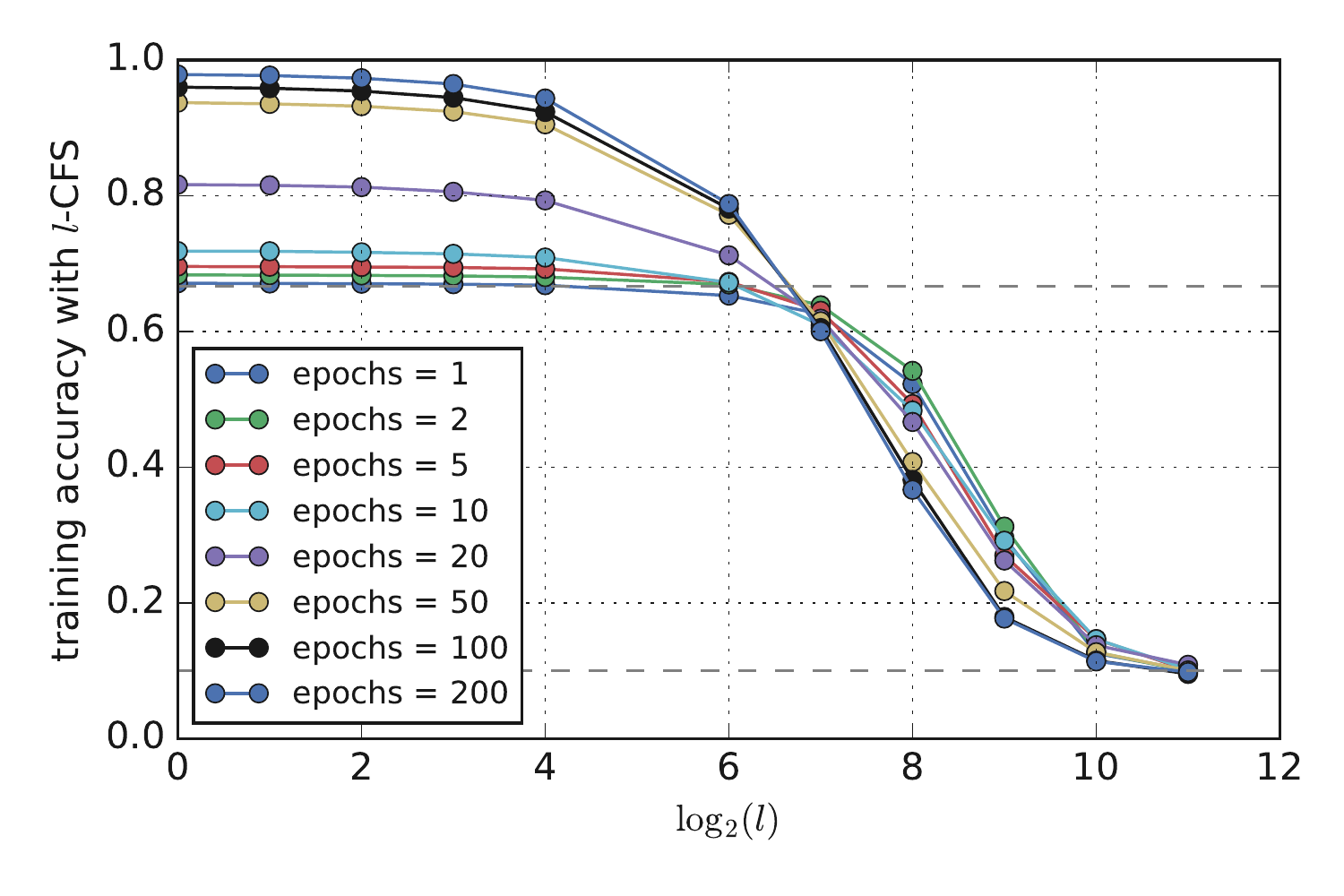}
    \caption{}
\end{subfigure}
    \hspace{1cm}
~
\begin{subfigure}[b]{0.40\textwidth}
    \includegraphics[width=\textwidth]{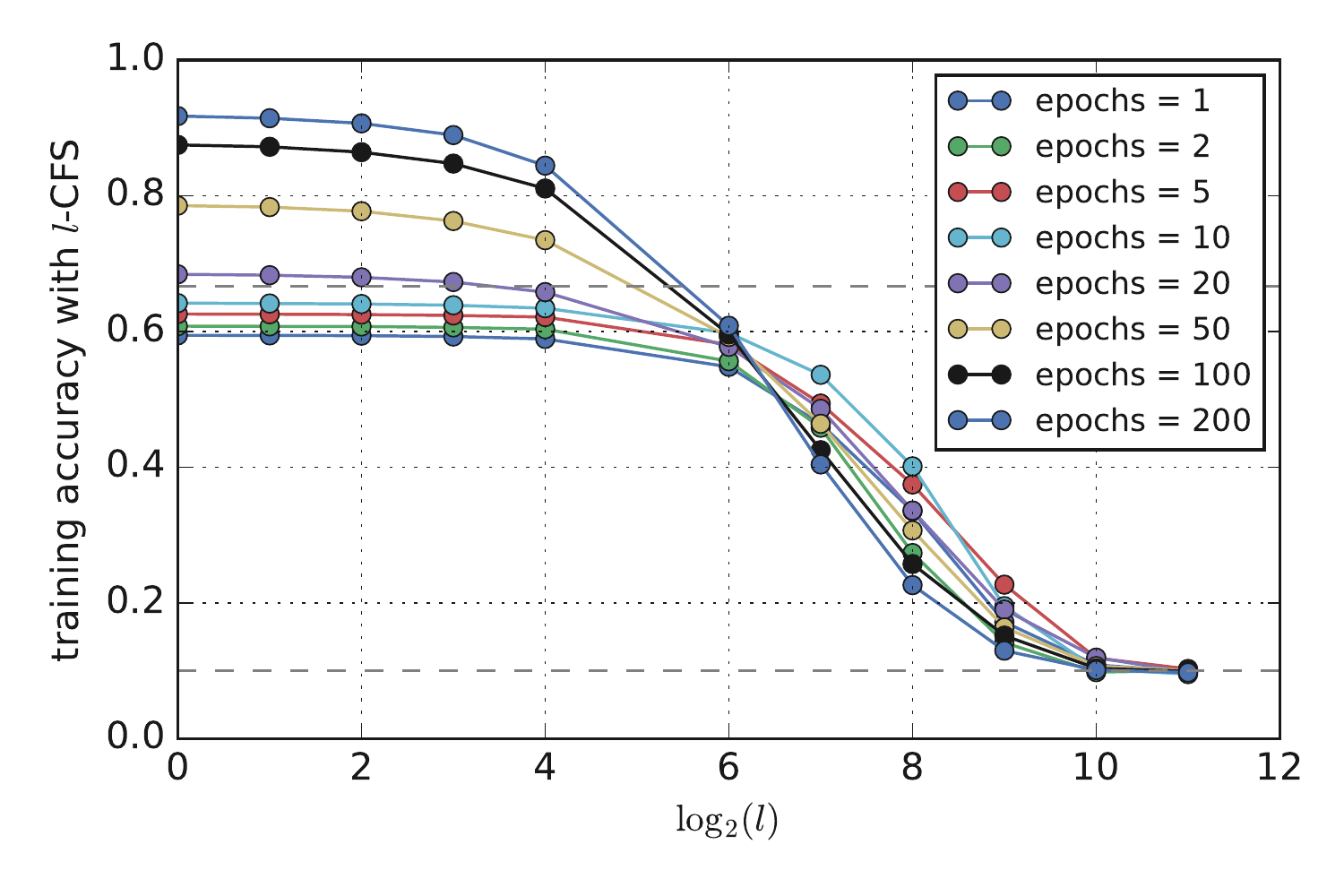}
    \caption{}
\end{subfigure}

\caption{
These plots show the result of repeating Expt. 7 when we corrupt only a third of the labels in (a) MNIST and in (b) Fashion-MNIST. In both cases, just as in Figure~\ref{fig:expt-7} we see that all the curves pass through a point close to the maximum achievable accuracy (around $2/3$). The reason for this is discussed in Section~\ref{sec:discussion} under Generalization in Deep Learning.
}

\label{fig:expt7third}
\end{center}
\end{figure*}

{\bf Expt. 2: Impact of Architecture.} There are many different ways in which a
neural network can be compiled down into logic gates. In Expt. 1, we made
certain architectural choices for the circuit, but what if we had chosen
differently? To evaluate that, here we replace the multipliers
used in Expt. 1 with array multipliers (i.e., multipliers based on the
elementary algorithm for multiplication). Figure~\ref{fig:expt-2} shows the
resultant CFS curves (with dashed lines) as well as the original curves (solid
lines) for reference.  The curves do not coincide indicating that the result of
CFS depends on the structure of the circuit and not just on the function
implemented by the circuit (since the function is the same in both cases).
However, for the same choice of architecture, we find that the falloff in CFS
curves are again indicative of the degree of overfit.

{\bf Expt. 3: Impact of Choice of Primitives.} Even at the lowest level of
abstraction, we can choose what primitives to work with. To see how this choice
impacts CFS curves, in this experiment we disallow Xor gates as primitives
(thus requiring that only And gates and inverters be used).
Figure~\ref{fig:expt-3} shows the resulting CFS curves (dashed) as well as the
originals (solid). Once again, we see that the curves do not coincide
indicating that the choice of primitives matters but for the same choice of
primitives, the falloff in CFS curves are again indicative of the degree of
overfit.

{\bf Expt. 4: Count of Rare Patterns.} Figure~\ref{fig:expt-4} shows for each
value of $l$, how many examples have {\em no} $l$-rare patterns, i.e., cannot
be possibly affected by $l$-CFS. We observe in particular, that for {\tt
nn-random}, there are 54,094 examples (about 90\% of the training set) that
have no 1-rare patterns. This is in sharp contrast to a simple lookup table
where every example would have a 1-rare pattern, and in fact slightly more than
{\tt nn-real-100}.
Comparing these curves of counts to the CFS curves in Figure~\ref{fig:expt-1}
indicates that perturbation is an important part of CFS and that rarity by
itself is a cruder measure of overfitting since it may not be observable.

{\bf Expt. 5: Random Forests.} Since CFS works on the circuit level, it can 
check random forests for overfit.  Two random forests were trained using version 0.19.1 of
Scikit-learn~\citep{scikit-learn}. Each forest has 10 trees and is trained
using the default settings, except for bootstrapping (to avoid non-uniform
weights during inference).
The first forest ({\tt rf-real}) was trained on MNIST whereas the second forest
({\tt rf-random}) was trained on MNIST with the output labels pseudo-randomly
permuted (as before with {\tt nn-random}).  Both forests reach perfect training
accuracy. {\tt rf-real} gets 95.58\% validation accuracy whereas as expected
{\tt rf-random} gets no better than chance. ({\tt rf-real} has about 14K nodes
per tree whereas {\tt rf-random} has about 70K nodes per tree.)

The forests are compiled down to circuits in a straightforward manner. Each
tree is compiled separately and produces 10 16-bit outputs (one per class). The
corresponding outputs are added across all 10 trees and the class output by the
forest corresponds to the class with the maximum value. Each internal node in a
tree maps to a multiplexer controlled by a 8-bit comparator to implement the
threshold, and each leaf node corresponds to 10 16-bit constants representing
the number of examples that occupy each class in that leaf (thus most entries
are zero). The circuit for {\tt rf-real} has about 700K nodes whereas {\tt
rf-random} has 3M nodes. Both are less than 250 logic levels deep. (These are much
smaller than the circuits for the neural networks.)

Figure~\ref{fig:expt-5} shows the CFS curves for the two random forests. Once
again, we see that the overfit model ({\tt rf-random}) degrades faster than the
model with better generalization ({\tt rf-real}) confirming that CFS is
effective even for models that are fundamentally different from neural
networks.
However, it is interesting to see that if we compare {\em across} model
families i.e. between the neural networks from Expt. 1 (repeated in
Figure~\ref{fig:expt-5} for convenience) and the random forests, CFS is not
effective at distinguishing overfit. In particular, {\tt nn-real-2} which is not
overfit degrades more rapidly that {\tt rf-random} which is highly overfit.
We discuss this in greater detail in Section~\ref{sec:discussion}.

{\bf Expt. 6: Blanket Noise.} CFS may be seen as adding a targeted noise.
Here, instead, we add blanket noise by simulating the training set while 
randomly flipping the node values with probability $p$. 
As $p$ varies from $2^{-30}$ to $2^{-5}$, the resulting {\em noise curves}
(Figure~\ref{fig:expt-6}) are similar to the CFS curves (Figure~\ref{fig:expt-5}).
However, the more overfit {\tt nn-real-100} does {\em not} fall faster
than {\tt nn-real-2}. With CFS, these curves are well separated, 
and the gap between neural nets and forests is much larger. 

{\bf Expt. 7: Sensitivity.} To better understand the sensitivity difference
between CFS and blanket noise, we trained 38 neural networks (with the same
architecture as before but different number of epochs) on MNIST with exactly one half of labels
randomized (so maximum accuracy possible is about 55\%). We show the CFS curves
and the noise curves for 8 representative networks in Figures~\ref{fig:expt-7}
and \ref{fig:expt-8} respectively. Note the crossover of the CFS curves that
indicates a larger falloff for overfit networks compared to the more uniform
degradation of the noise curves. It is fascinating that all the CFS
curves cross over at a single point with an accuracy of about 50\%. This is discussed 
in Section~\ref{sec:discussion}.

{\bf Expt. 8: Composite CFS.} For completeness, Figure~\ref{fig:expt-composite} shows the results of running Composite CFS instead of Simple CFS in the setup of Expt. 7. We see that these curves are very similar to those of Simple CFS in Figure~\ref{fig:expt-7}.

\section{Discussion}
\label{sec:discussion}

{\bf Structure Dependence}. 
Expt. 2 and 3 show that the results of Simple CFS
depend not just on the function but on the structure of the circuit. (Other CFS
variants we investigated show this behavior as well.) 
A small example provides some insight.  Consider the Boolean function $f(a, b,
c) = a$ evaluated on the training set comprising the full Boolean cube (i.e.,
all 8 combinations of $a, b, c \in \{0, 1\}$).  In addition to the direct
implementation, $f$ can also be implemented (redundantly) as
$a \cdot b \cdot c + a \cdot \lnot b \cdot c + a \cdot b \cdot \lnot c + a
\cdot \lnot b \cdot \lnot c$.
It is easy to see that under 1-CFS, the direct implementation is unchanged
(there are no 1-rare patterns) but the redundant implementation maps to
constant 0 (the output of each conjunction is 1 only once).

Although this is not a problem when the compilation process can be controlled,
this is bad news in the adversarial setup.  A good model with a poor
implementation may show steeper degradation under CFS than a more overfit model
with better implementation (c.f. {\tt nn-real-2} with array v/s {\tt
nn-real-100} without Xors at $l = 256$).
Ideally, we would like to find a variant of CFS that does not depend on
structure but only on the function.\footnote{In principle, one could canonize
the circuit structure before applying CFS, say by building Reduced Order Binary Decision Diagrams (ROBDDs), but that
would be computationally prohibitive. Alternatively, one could lightly optimize
the circuit before CFS but that may not be enough.}  
In the absence of that ideal, we view the structure of the circuit as a
certificate of how well the dataset is learnt, and make it Merlin's
responsibility to find and present the most convincing structure. 
From this perspective, in the above example, the direct implementation of $f$
(which is not impacted by 1-CFS) is more convincing than the redundant
implementation of $f$ (which is severely impacted). 

{\bf Adversarial Attack on CFS.} 
Based on the discussion above it is easy to
design a way to arbitrarily degrade the performance of a circuit
under CFS. But is the opposite possible? Can Merlin fit an arbitrary function on the
inputs but compile it down to a circuit which does not degrade under CFS?
Expt. 5. offers a clue. The overfit model {\tt rf-random} fitted on
random labels falls off more slowly than {\tt nn-real-2} which generalizes well.
What is going on? The short answer is that although each tree in {\tt
rf-random} is extremely overfit with most leaves containing only a single
example, the circuit nodes have few rare patterns due to the
observability don't cares introduced by the multiplexers.

Again a simple example is instructive. Let $f$ be the parity function on $n$
bits, i.e., $f(x_1, x_2,\dots,x_n) = x_1 \oplus x_2 \dots \oplus x_n.$ Consider
an tree implementation of this function obtained using Shannon decomposition
which has a multiplexer at the top controlled by $x_1$ and with $1
\oplus x_2 \oplus x_3 \dots \oplus x_n$ and $x_2 \oplus x_3 \dots \oplus x_n$ as its data inputs. 
If the training set is the full Boolean cube, i.e., $\{0, 1\}^n$ it is easy to
see that there are no $l$-rare signals for $l << n$ since each input to the
multiplexer is balanced, i.e., has equal number of 0s and 1s. 
Since Shannon decomposition is recursive, a similar argument holds for the
lower levels of the tree until we get to the leaves which are constant 0 and
constant 1 which though unbalanced have no rare patterns.

This example suggests a way to break CFS. 
If Merlin wanted to build a lookup table, but prevent Arthur from identifying it as such (by applying say 1-CFS), instead of building the circuit in Figure~\ref{fig:lut_circuit}, Merlin could build a decision (i.e., multiplexer) tree by recursively
splitting on one variable (i.e., one component of $x$) at a time by picking a component that is balanced (i.e., takes on 0 and 1 values roughly equally often) and leads to both branches also being balanced (i.e., have no class with either too few or too many examples, although having no examples of a class or having only examples of one class is fine and expected at the leaf nodes).
In other words, Merlin could use a procedure similar to 
the usual decision tree construction procedures, but with the important difference of favoring balanced splits instead of unbalanced splits. (The unbalanced
splits are likely why CFS is effective to the extent it is on random forests.)

Finally, as suggested by a reviewer of this paper, it would be interesting to extend this idea (ultimately based on Shannon decomposition and co-factoring) to an algorithm that can amplify the count of rare patterns in a given circuit without changing its functionality.

{\bf Comparison with Blanket Noise.} 
Based on Expt. 6 and 7, we believe that
blanket noise is less sensitive than CFS. Our results here add to the extensive
literature on noise, generalization and fault tolerance in neural networks
(e.g., see~\citet{Bernier01} and the references therein) by extending them to
the circuit level (where distinctions between activation or weight noise, or
additive or multiplicative noise disappear) and to other model families such as
random forests.
Furthermore, Expt. 6 presents a direct comparison of the fault-tolerance of
neural nets and forests where forests are seen to be about 1000x more
fault-tolerant to bit flips. This is likely due to the redundancy from ensembling. It also
suggests that noise-based intrinsic methods could be easily fooled by an
adversary by adding redundancy.

{\bf Generalization in Deep Learning.} 
Why do neural nets trained with SGD generalize when they have sufficient
(effective) capacity to memorize their training set? This is an open research
question~\citep{Zhang17, Arpit17, Bartlett17, Arora18, Neyshabur18}.
Expt 5. shows that this question is not limited to nets---the same could be asked for
random forests as well. One (informal) answer for forests is that decision tree
construction procedures look for common patterns between examples. When
examples share commonality, they are combined into common leaf nodes and the
model generalizes, whereas when there is little commonality, each example is
its own leaf (so the training set is fit well) but the model fails to
generalize.
Could the same thing be going on with SGD and networks? CFS on {\tt nn-random}
and Expt 4. provide {\em direct} evidence that even on random data, nets do not
``brute-force memorize'' but identify common patterns in the data (a question
raised in \citet{Zhang17} and discussed in ~\citet{Arpit17}).

In this context, it is interesting to study why the CFS curves for the
different networks in Expt 7. intersect at a common point corresponding approximately to the
achievable accuracy. This holds for other amounts of label randomization e.g., see Figure~\ref{fig:expt7third} for the CFS curves when one third of the labels are corrupted.
Roughly half of the examples are easy since they have correct labels and are learnt in the
first few epochs.  The remaining examples with corrupt labels are harder and
learnt only in later epochs by the models that are trained longer. With CFS,
the accuracy of those models breaks down earlier since the hard examples have
more rare patterns than the easy examples, and the accuracy on the easy
examples thus forms a limiting curve for all models. 
This provides more (and direct) evidence for
the claim in \citet[\S 1]{Arpit17} that ``SGD learns simpler patterns first before
memorizing.'' Furthermore, we could identify ``simpler patterns'' as examples that
have fewer rare patterns and ``memorizing'' as what is required for examples
that have more rare patterns. Thus, learning simpler patterns and memorization are not
fundamentally different but lie at two ends of a spectrum.

\balance

{\bf Related Work.} One may be tempted to view margin as an intrinsic measure
to estimate the generalization of a model. However, when we have models with
intermediate representations, the notion of margin by itself is not adequate
since an adversary can overfit to a favorable intermediate representation that
is easily linearly separated (but otherwise arbitrary).
However, recent work in this area (e.g., \citet{Bartlett17}) has focused on
margins normalized by spectral complexity (i.e., a measure related to the
Lipschitz constant of the network) and in that case the above argument does not
obviously apply. We have not studied if normalized margin can be exploited by an
adversary.
Similarly, most measures based on the shallowness of minima are not adequate
since they are not scale-invariant~\cite{Dinh17} and we have not
investigated if more recent work on scale-invariant
measures~\citep{Rangamani19} can be exploited.
In comparison to these and other approaches for neural
networks~\citep{Arora18, Neyshabur18}, CFS is fundamentally more discrete, which
makes it applicable to a larger class of models. However, in contrast to the
other approaches, we do not have any theoretical bounds yet while our
results indicate that without further refinements to CFS itself, any generalization
bounds from $l$-CFS would likely be vacuous in practice.

\section{Conclusion and Future Work}

Our main result is that CFS based on adding small amounts of targeted noise at the logic
circuit level can detect overfit. This is remarkable because at this level
of representation we have lost most aspects of the structure of the model, such as
the distinction between weights and activations, or even whether the model is a
neural network, a random forest, or a lookup table.  Furthermore, variations
such as perturbing only rare patterns in single signals or across the fanins of
a gate 
lead to qualitatively similar results and there are variants (such as Simple
CFS) that are naturally free of hyper-parameters.

By studying rare patterns, we find that SGD does not lead to ``brute force''
memorization, but finds common patterns (whether training is done with randomized
labels or actual labels), and neural networks are not unlike forests in this
regard. By adding blanket noise, random forests are found to be about 1000x 
more resilient to noise than neural networks which could be
useful when implementing machine learning systems with unreliable low level
components. 

There are several directions for future work. We analyze flat circuits, 
but with a clever implementation that constructs the circuit on-the-fly 
from a higher level specification, the computations can scale to larger models.  
We could also apply CFS at higher levels of abstraction (perhaps as part of the model
evaluation process in frameworks, such as Scikit Learn and Tensorflow Estimators)
though at that level there are more degrees of freedom in the implementation
(e.g., what kind of noise to add). 

The notion of rarity considered in this work may be regarded as a local notion, since we count the patterns at a signal (or a group of signals in the case of Composite CFS) in isolation. 
It is possible to extend this notion to a global notion of rarity by propagating occurrence counts along with signal values during simulation. 
In the corresponding CFS, a pattern is perturbed when its global rarity drops below the given threshold. Preliminary experiments show that this more stringent notion of rarity is more effective in detecting overfit in the case of random forests, and may offer a way to partially mitigate the 
adversarial attack on CFS outlined in the previous section.

Finally, based on insights from our analysis of Simple CFS, we would like to
continue the search for an intrinsic method that does not depend on the model structure
and is adversarially robust, or to show that such a method does not exist, 
even for learning tasks of practical interest.
As a generalization of this idea, it is interesting to contemplate learning
algorithms that produce certificates of generalization, much like a
Boolean satisfiability solver can produce a certificate of satisfiability or of
unsatisfiability.

{\bf Postscript: The Theory of Coherent Gradients.} The observations in this paper from applying CFS to neural networks\,---\,particularly, the absence of ``brute force'' memorization with random labels, and the existence of easy and hard examples as discussed in Section~\ref{sec:discussion}\,---\,inspired the development of a theory called Coherent Gradients (CG) that provides a simple and intuitive explanation of generalization in networks trained with (stochastic) gradient descent.

The key idea in CG is as follows. 
Since the overall gradient $g$ is the average of the gradients of the individual training examples, there may be certain components (directions) of $g$ which are significantly stronger than other components due to the gradients of multiple examples being in agreement on those components, and thereby reinforcing each other. 
Since the changes to the trainable parameters of the network are proportional to the gradient, the biggest changes to the parameters are biased to benefit multiple examples, and therefore likely to generalize well (based on a stability argument).
However, if there are no such strong components, i.e., all the per-example gradients are roughly orthogonal, then each example is fitted independently, and this corresponds to memorization (and poor generalization, again from a stability argument).
Thus CG provides an uniform explanation of both generalization and memorization in neural networks.     

Please see \citet{Chatterjee20} and \citet{Zielinski20} for a more detailed development of this idea, including an analysis of easy and hard examples from this perspective, and a natural modification to gradient descent that significantly reduces overfitting by suppressing those gradient components that are not common to many examples.

\section*{Acknowledgments}

The first author thanks Michele Covell, Ali Rahimi, Alex Alemi, Shumeet Baluja, Sergey Ioffe, Tomas Izo, Shankar Krishnan, Rahul Sukthankar, and Jay Yagnik for helpful discussions. The second author was supported in part by SRC Contract 2867.001.


\bibliography{paper}
\bibliographystyle{icml2020}

\end{document}